# Joint angle based learning to refine kinematic human pose estimation

Chang Peng [1], Yifei Zhou [1], Haoqiang Ren [1], Shiqing Huang [2], Chuangye Chen [3], Jianming Yang [3], Bao Yang [1*], Huifeng Xi [2*], Zhenyu Jiang [1*]

[1] Department of Engineering Mechanics, School of Civil Engineering and Transportation, South China University of Technology, Guangzhou 510640, China

[2] School of Mechanics and Construction Engineering, Jinan University, Guangzhou 510632, China.

[3] Guangdong Provincial Key Laboratory of Speed Capability, School of Physical Education, Jinan University, Guangzhou 510632, China.

*Corresponding authors: byang20210415@scut.edu.cn (B Yang), xihuifeng@jnu.edu.cn (H Xi), zhenyujiang@scut.edu.cn (Z Jiang)

## Abstract

Marker-free human pose estimation (HPE) has found increasing applications in various fields. Current HPE suffers from occasional errors in keypoint recognition and random fluctuation in keypoint trajectories when analyzing kinematic human poses. The performance of existing deep learning-based models for HPE refinement is considerably limited by inaccurate training datasets in which the keypoints are manually annotated. This paper proposed a novel method to overcome the difficulty, in which the key techniques include: (i) A robust joint angle-based description of kinematic human poses; (ii) Approximating temporal variation of joint angles using high order Fourier series to get reliable "ground truth"; (iii) A bidirectional recurrent network is designed as a post-processing module to refine the estimation of single image-based HPE models. Trained with the high-quality dataset constructed using our method, the network demonstrates outstanding performance to correct wrongly recognized joints and smooth their spatiotemporal trajectories. Tests show that joint angle-based refinement (JAR) outperforms the state-of-the-art HPE refinement network in challenging cases like figure skating and breaking. JAR also demonstrates great potential to rectify existing datasets.

## 1. Introduction

Computer vision based human pose estimation (HPE) has been widely adopted as a powerful tool to determine the configuration of human body from images and videos. This technology has found increasing applications across various fields such as human-computer interaction, motion analysis, augmented or virtual reality, and healthcare [1]. HPE also gains the success in sports, serving as coach systems which offer indispensable approach to analyze and quantify athletic performance [2–6], as well as referee systems which help to make more accurate ranking [7].

HPE is expected to reproduce high-resolution model of moving human body [8], which means accurate locations of the keypoints of body as well as their trajectories. Deep learning based HPE models, experienced swift progress over the past decade, achieve impressive capability to catch human poses from images. However, there remain some challenges:

- Incorrect recognition of keypoints. The complexity and diversity of human poses in various activities bring considerable ambiguities to HPE. In addition, visual occlusion and distracting background are prone to induce erroneous recognition of some keypoints. Those outliers, even appear in a small number of frames, may substantially disturb the analysis of human motion, e.g., calculation of velocity and acceleration, as well as inverse dynamics-based estimation of force and moment.
- Persistent jitters in spatiotemporal trajectories of keypoints. The issue is ascribed to the inconsistent localization of keypoints among the frames which record continuous human motion. These fluctuations also complicate the subsequent analysis of human motion, reducing the reliability and applicability of HPE technology.

Deep learning based HPE, as a typical data-driven task, relies highly on training datasets. COCO-WholeBody [9] is one of the most popular HPE datasets, which contains over 200,000 images and 250,000 instances with around 133 annotated keypoints on human body. Inconsistence of annotation is inevitably existing in this dataset due to the diversity of annotators and their criteria, which may result in non-negligible deviation of keypoint localization among a series of images recording human motion. The problem could be further aggravated in the cases of derivative calculation for analysis of speed and acceleration. Another limitation of single image dataset, when being used as benchmark for motion analysis, is that it tends to make HPE models excessively emphasize the Euclidean distance between the locations of recognized keypoints and the annotated ones, while ignoring the continuity of their spatiotemporal trajectories [10–12]. In consequence, some HPE models reach high scores in this kind of metrics such as percentage of correct parts [11] or percentage of correct keypoints [12], whereas suffering considerably the jitters in output trajectories. Video datasets (e.g., PoseTrack [13] and HiEve [14]) may guide the models to take the continuity of body motion into account. However, the variety of activities included in current video datasets is far from satisfactory [13–17]. These datasets are also bothered by the inconsistency of manual annotation, especially in cases of high-amplitude and fast-changing motions. Thus, it remains as a common procedure when dealing with sports that the video is resolved into images, followed by the processing with a single image-based HPE model [6,7,18–22].

To improve the performance of HPE refinement for kinematic poses, we develop a joint angle-based refinement (JAR), which can provide reliable ground truth of smooth joint trajectories for network training and enhance the inter-frame consistence of human pose estimation. The main contributions of this paper can be summarized as follows:

- A joint angle-based model is proposed to describe kinematic human poses. The description is resistant to the interferences caused by the changes in perspective and distance of observation during recording of human activities. Combined with the rule of human motion kinematics and temporal invariant of limb length in reconstruction of keypoint location, the accuracy and robustness of human body recognition are significantly ameliorated.

- The variation of joint angles in human activities is approximated using the high order Fourier series. By fitting the parameters in series according to the open datasets containing joint angle evolution in common human activities and sports, the model can generate the dataset with spatiotemporal consistence and continuity of joint angles for training of refinement network.
- A bi-directional gated recurrent unit network with attention mechanism (BiGRU-Attention) is designed to refine the outputs of single image-based HPE models. Trained with our dataset, BiGRU-Attention demonstrates outstanding performance to eliminate outliers and jitters in keypoint tracing. Some challenging cases show that our model outperforms SmoothNet.
- JAR works not only for motion capture and analysis. It can also be used to rectify the existing video datasets, minimizing the adverse effects of inaccurate and inconsistent annotation.

## 2. Related work

The research effort dedicated to improving the accuracy and stability of HPE for human motion can be categorized into two directions: (i) Thorough exploration of information in single image; (ii) Efficient integration of inter-frame relation.

### 2.1. Single Image-based HPE

Deep learning has significantly enhanced the capability of HPE in extracting information from single images. DeepPose [23] firstly employs convolutional neural networks (CNNs) for HPE, utilized AlexNet as a backbone architecture to estimate coordinates keypoints. Joint-CNN-MRF [24], proposed in the same year, introduced a hybrid architecture combining CNNs with Markov random fields to extract spatial geometric relationships among joints, effectively reducing false-positive estimation. Notably, it pioneers heatmap-based localization, where keypoints are determined by the centroids of pixel-wise probability maps, as a paradigm widely adopted in subsequent studies [5]. Convolutional Pose Machines [25] further improves the accuracy through a multi-stage design that iteratively refines spatial and image features. Stacked Hourglass Networks [26] incorporates U-Net hierarchical structure into HPE. It enables multi-scale feature fusion via stacked hourglass modules. This architecture inspires variants like PyraNet [27] and MSS-net [28], which enhance cross-scale feature matching.

HRNet proposed by Wang et al. [29] in 2019 represents a milestone in HPE. This model innovatively introduces parallel architecture connecting high-resolution and low-resolution subnetworks, enabling the maintenance of high-resolution feature representations throughout the entire process, thereby achieving more accurate heatmap estimation. It has become one of the most widely used models in single image-based HPE. In recent years, attention mechanisms have been integrated into HPE. Soft-gated skip connections incorporate channel-wise attention during feature extraction [30]. Transformer encoders were employed to enhance the features extracted by CNNs. ViTPose [31] demonstrates the great potential of Vision Transformer for HPE. Its extension, ViTPose++ [32], represents the top tier in single image-based HPE.

### 2.2. Video-based HPE

Since 2018, researchers have begun to explore inter-frame correlations in videos to improve the stability and accuracy of HPE. The LSTM Pose Machines [33] extends the Convolutional Pose Machines [25] by incorporating Long Short-Term Memory (LSTM) units to capture temporal dependencies between frames. Similarly, Temporal Feature Correlation HPE [34] highlights the importance of exploiting temporal information for video-based pose estimation. This model extracts features of previous frames to guide feature search in current frames, improving robustness in estimating high-amplitude activities. UniPose-LSTM [35] achieves notable performance by fusing the decoded heatmaps of current frame with those generated by an LSTM module from previous frames. Attention mechanisms, renowned for their ability to model long-range dependencies, have also been integrated into video-based HPE frameworks [36–38]. For example, SLT-Pose [36] takes HRNet as a backbone to extract coarse pose features and constructs sequential keypoint information across frames. A cross-frame temporal learning module with attention mechanisms is used to enhance the information interaction between the target frame and local sequences. However, the generalization capability of current video-based HPE models seems inferior to that of their single image-based counterparts. This performance gap may be attributed to the substantially limited scale and diversity of samples in existing video datasets, compared with single image datasets [9,10,39–41].

### 2.3. HPE Refinement

To effectively extract temporal information while preserving the advantages of single image-based HPE models, researchers try to refine keypoint trajectories obtained by single image-based HPE. Traditional bandpass filters can remove noise within specific frequency. However, human motion spans a wide frequency spectrum, making it challenging to define universal thresholds for various noise patterns. Modern filtering techniques, such as Kalman filters, perform well under specific conditions but rely heavily on reasonable parameter setting. They are also struggling to fit high-amplitude and fast-changing motions, often resulting in over-smoothing or drift.

Learning-to-Refine-HPE [42] introduced deep learning for HPE refinement. PoseFix [43] extends this work by analyzing the error distribution of HPE outputs and incorporating error statistics as the prior knowledge. Additionally, PoseFix adopts coarse-to-fine architecture, trying to make refined poses satisfy the biomechanical constraints. While these designs achieve improvements in single-frame correction, the lack of temporal modeling limits their ability to mitigate inter-frame jitters. To overcome this limitation, subsequent studies improved spatiotemporal feature extraction during decoding phase. Pose Temporal Merger [44] integrates auxiliary information from adjacent frames during heatmap generation, providing search ranges for keypoint localization in current frame. This approach improves the accuracy in scenarios with motion blur, defocus, or occlusion. SynSP [45] proposes a multi-view optimization method, where pose sequences from multiple perspectives are independently encoded and fused according to pose similarity-based weighting. This method effectively corrects erroneous recognition by emphasizing consistent poses across views. In these studies, pose estimation is tightly coupled with refinement within a unified architecture. Such strong interdependence may lead to performance degradation when working with different HPE models. To address this issue, decoupled strategies were proposed that take single image-based HPE results as the inputs to refinement module. Temporal PoseNet [46] adopts a two-stage architecture: the single-image HPE model processes each video frame, and the output keypoint sequences are refined by Temporal Convolutional Networks (TCNs). It generates smooth temporal keypoint trajectories, bridging gaps in visibility. SmoothNet [47], regarded as the state-of-the-art refinement framework, also follows the two-stage pipeline, appending a specific temporal smoothing network to single image pose estimators. The framework effectively utilizes the sample diversity of single image datasets while incorporating temporal modeling, achieving significant improvements in human motion smoothing. However, performance of SmoothNet in complex motion scenarios remains limited, with refined results frequently occurring noticeable position drift (examples are given in Section 5).

## 3. Modeling

### 3.1. Joint angle-based description of Human Pose

Fig. 1 illustrates the processing of the joint angle-based model, which is expected to mitigate the issue of inter-frame inconsistence arising from changes in shooting distance and viewing angle. The coordinates of 13 keypoints on human body are extracted from each frame using single image-based

HPE, including nose, shoulders (left and right), elbows (left and right), wrists (left and right), hips (left and right), knees (left and right), ankles (left and right). The coordinates of these keypoints ($p$) form the position matrix $\mathbf{M}_1$:

$$\mathbf{M}_1 = \begin{bmatrix} p_1^1 & p_1^2 & \cdots & p_1^{13} \\ p_2^1 & p_2^2 & \cdots & p_2^{13} \\ \cdots & \cdots & \cdots & \cdots \\ p_n^1 & p_n^2 & \cdots & p_n^{13} \end{bmatrix} \tag{1}$$

Each row of $\mathbf{M}_1$ comprises the spatial coordinates of 13 keypoints in the $i$-th frame (where $i$ is the subscription of $p$). Using the keypoint "nose" (which exhibits the highest recognition accuracy and stability) as the base point $p_i^1 = (x_i^1, y_i^1)$. Every three adjacent keypoints form two vectors (such as $\overrightarrow{OA}$ and $\overrightarrow{OB}$ in Fig. 1). The angle of the two vectors is treated as the joint angle $\theta$ at keypoint $O$. It is calculated through arctangent function:

$$\theta_i^O = \operatorname{arctan2}(\overrightarrow{OA}, \overrightarrow{OB}) = \arctan\left(\frac{y_i^B - y_i^A}{x_i^B - x_i^A}\right) \tag{2}$$

where $(x_i^A, y_i^A)$ and $(x_i^B, y_i^B)$ represent the pixel coordinates of keypoint $A$ and keypoint $B$, respectively. By this way, 12 joint angles are derived from the coordinates of 13 keypoints, which are assembled into matrix $\mathbf{M}_2$:

$$\mathbf{M}_2 = \begin{bmatrix} \theta_1^1 & \theta_1^2 & \cdots & \theta_1^{12} \\ \theta_2^1 & \theta_2^2 & \cdots & \theta_2^{12} \\ \cdots & \cdots & \cdots & \cdots \\ \theta_n^1 & \theta_n^2 & \cdots & \theta_n^{12} \end{bmatrix} \tag{3}$$

The coordinates of the base point along $X$-axis and $Y$-axis are smoothed in temporal domain using a Savitzky-Golay filter. Local quadratic polynomial fitting is performed on the original coordinate system within a sliding window. The coordinate of the base point along *X*-axis or *Y*-axis is supposed to follow the expression:

$$\hat{x}_i^1 = c_0 + c_1 i + c_2 i^2 \tag{4}$$

where $c_0$, $c_1$ and $c_2$ denote constant coefficients. They can be estimated by minimizing the function:

$$\min \sum_{k=i-w}^{i+w} \left\| x_k^1 - (c_0 + c_1 k + c_2 k^2) \right\|^2 \tag{5}$$

The size of sliding window $w$ is set as 50 frames in this study, which is half of image collection rate (about 100 fps) in experiments. Smoothing the trajectory of base points yields a stable sequence $S_B = [\hat{p}_1^1, \hat{p}_2^1, ..., \hat{p}_n^1]$, where $\hat{p}_i^1 = (\hat{x}_i^1, \hat{y}_i^1)$. Then the coordinates of all the other keypoints can be determined, combining with the base point positions and the limb lengths.

To estimate the lengths of limbs, two inherent physiological constraints derived from human biomechanics are imposed: (i) Proportions between limb lengths remain approximately consistent; (ii) Foreshortening-induced inter-frame variations of limb lengths could be negligible. Based on the two constraints, the objective function in optimization is designed as follows:

$$Loss = \sum_{i,j}^{12} \sum_{n=1}^{N} \left( \frac{\hat{L}_n^i}{\hat{L}_n^j} - R_{i,j} \right)^2 + \sum_{i=1}^{12} \sum_{n=2}^{N} \left( \hat{L}_n^i - \hat{L}_{n-1}^i \right)^2 \tag{6}$$

where $L_n^i, L_n^j$ represent the length of the *i*-th and *j*-th limbs in the *n*-th frame, $\hat{L}_n^i, \hat{L}_n^j$ indicate the lengths of corresponding limbs after optimization. $R_{i,j}$ is the ratio of $L_n^i, L_n^j$. With a total of 12 limbs and *N* frames, the limb length matrix $\mathbf{M}_L$ for all frames is derived through iterative optimization using the trust region algorithm.

$$\mathbf{M}_L = \begin{bmatrix} \hat{L}_1^1 & \hat{L}_1^2 & \cdots & \hat{L}_1^{12} \\ \hat{L}_2^1 & \hat{L}_2^2 & \cdots & \hat{L}_2^{12} \\ \cdots & \cdots & \cdots & \cdots \\ \hat{L}_n^1 & \hat{L}_n^2 & \cdots & \hat{L}_n^{12} \end{bmatrix} \tag{7}$$

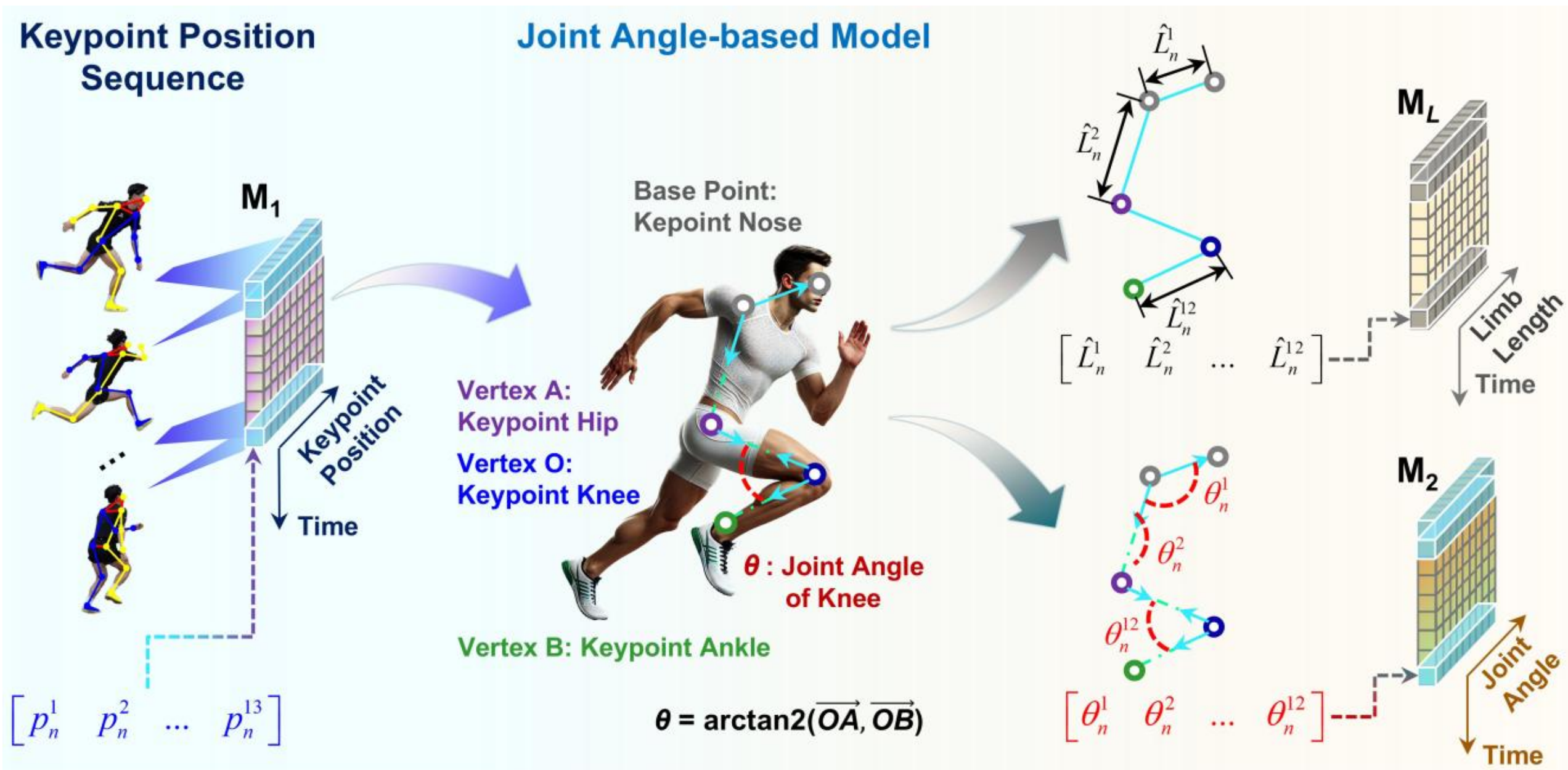

Fig. 1. Illustration of the joint angle-based model of human pose. Keypoint position sequences are stored in matrix $\mathbf{M}_1$, where each row represents the positions of 13 keypoints in a specific frame. Each row in $\mathbf{M}_1$ is used to derive a joint angle sequence through vector operation. All joint angle sequences are then temporally stacked to construct matrix $\mathbf{M}_2$. Additionally, the limb length sequence is calculated by the Euclidean distances between adjacent joint positions, which are then temporally stacked to construct matrix $\mathbf{M}_L$.

### 3.2. Fourier Series-based approximation of joint angle variation

Biomechanical and anatomical studies demonstrate that joint motion is constrained by osseous structure, tendon elasticity, and ligament tension. Consequently, temporal variations of joint angles follow specific physiological patterns and exhibit periodic characteristics in many activities. Open datasets indicate that the motion of three major lower-limb joints during running, i.e., hip, knee, and ankle, can be regarded as periodic movement [48–53]. Fig. 2(a) illustrates six phases of a running gait cycle. To accurately characterize these phases and ensure that inter-phase transitions satisfy biomechanical principles, an 8th-order Fourier series is used to approximate joint angle variations:

$$\theta(m) = a_0 + \sum_{k=1}^{8} \left[ a_k \cos\left(\frac{2\pi km}{T}\right) + b_k \sin\left(\frac{2\pi km}{T}\right) \right] \tag{8}$$

where $\theta(m)$ denotes the joint angle as a function of frame index $m$ in a cycle. $T$ represents the period of motion. $a_0$ is a constant corresponding to the mean value of joint angle. $a_k$ and $b_k$ are the coefficients of the trigonometric series, capturing the amplitude of different frequency components. These coefficients can be estimated via least squares regression on open datasets.

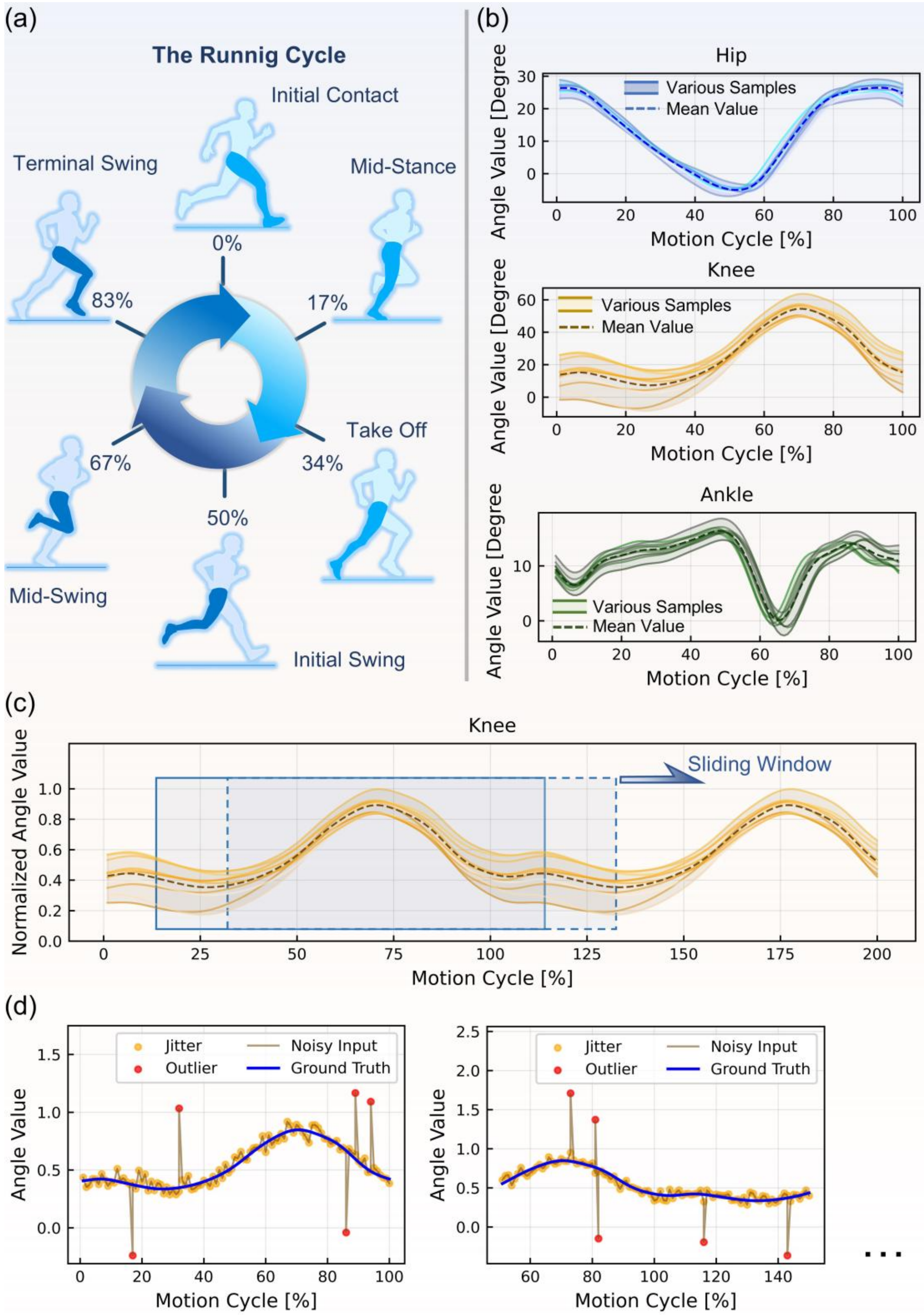

Fig. 2. Illustration of generating a joint angle dataset based on Fourier series. (a) Decomposition of the motion cycle. (b) Adjustment of Fourier series parameters to generate a series of angle variation curves for different joints within one motion cycle. (c) Extracting fixed-length segments of "ground truth" from two consecutive motion cycles using a sliding window. (d) Addition of small-amplitude jitters and large-amplitude outliers into the "ground truth" samples to generate noisy inputs.

### 3.3. Generation of training dataset

Generation of the training and testing datasets of joint angle variation can be summarized as a three-step procedure:

Step 1: Adjust the parameters of Fourier series ($a_0$, $a_1$, $b_1$, $a_2$, $b_2$, and $T$). The coefficient $a_0$ governs the joint angle baseline, while the other coefficients ($a_1$, $b_1$, $a_2$, $b_2$) and period $T$ modulate the amplitude and phase variations of the motion. Various combinations of these parameters can simulate inter-individual differences, thereby generating diverse training samples. Fig. 2(b) gives three examples of lower-limb joint angle curves (at hip, knee, and ankle) synthesized using different Fourier series configurations. The variations are controlled within a reasonable range.

Step 2: Slide window to segment samples of sequence. Considering that the model lacks prior knowledge about the exact temporal position of input data with arbitrary length during inference, a sliding window sampling strategy is used to create samples in dataset. Fig. 2(c) shows the sampling procedure, in which each motion cycle is discretized into 100 frames, and two consecutive cycles are selected to form a 200-frame sequence. A 100-frame sliding window with a stride of 1 frame is applied to the sequence, ensuring good coverage of all motion phases. The study ultimately generates 512,000 segments serving as the "ground truth" for training set, and 128,000 segments for testing set.

Step 3: Add noise and outliers into datasets. To improve the tolerance of the HPE refinement to anomalous perturbations, two types of synthetic disturbance signals are added to the samples, as shown in Fig. 2(d). The small-amplitude jitters are approximated as Gaussian white noise with the three times standard deviation ($3\sigma$) less than 45°. The noise amplitude matches the slight deviations that exist in the outputs of single-image HPE model. 5% of the total frames in samples are randomly selected as outliers containing erroneous results. In each outlier, Gaussian white noise with $3\sigma$ up to 135° is superimposed to a randomly chosen joint angle. A small number of secondary abnormal frames are placed before and after each outlier. The overall number follows Gaussian distribution with the average of zero and $3\sigma < 6$. It simulates the phenomenon of continuous recognition error observed in single image-based HPE.

## 4. Smoothing of human pose sequence

Fig. 3(a) illustrates the processing of JAR, which can be divided into three stages. In stage 1, single image-based HPE is employed to obtain pose estimation in each frame of human activities. The results (coordinates of keypoints) are transformed into joint angles. In stage 2, the joint angle sequence is smoothed using a deep-learning model (BiGRU-Attention) trained with the dataset described in Section 3.3. In stage 3, keypoint coordinates are reconstructed according to the smoothed joint angle sequence $\mathbf{M}_3$, in which $S_B$ is the first row, as described in Section 3.1.

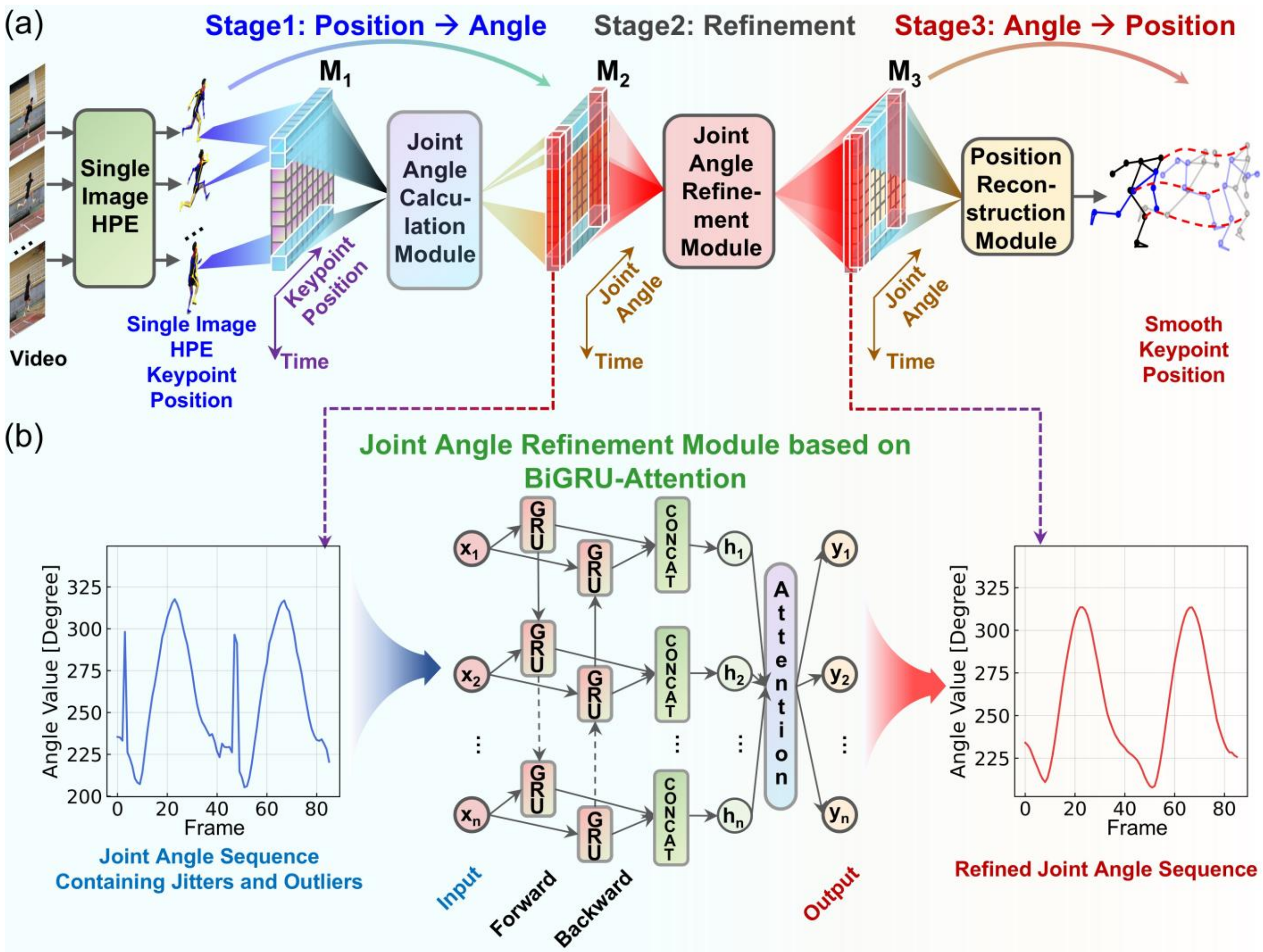

Fig. 3. The framework of kinematic human pose refinement. (a) In stage1, each frame of a video clip is fed into single image-based HPE model for pose estimation. The coordinates of keypoints output containing jitters and outliers, forms matrix $\mathbf{M}_1$. Each row of $\mathbf{M}_1$ is fed into the Joint Angle Calculation Module. The obtained joint angles at all moments are arranged to form matrix $\mathbf{M}_2$. In stage2, each column of $\mathbf{M}_2$ (i.e., the angle sequence of a specific joint over time) is fed into the Joint Angle Refinement Module. BiGRU-Attention model is employed to smooth the angle sequences, yielding matrix $\mathbf{M}_3$. In stage 3, each row of $\mathbf{M}_3$ is input into the Position Reconstruction Module to reconstruct keypoint positions. (b) Joint Angle Refinement module based on BiGRU-Attention.

### 4.1. Bidirectional recurrent network with attention mechanism

Fig. 3(b) shows the architecture of BiGRU-Attention model. The GRU layers are designed to retain relevant information while discarding irrelevant details. Although, in theory, increasing the number of GRU layers can improve the ability to capture nonlinear patterns, an excessive number of layers may lead to overfitting and significantly increase computational complexity. Thus, the model uses a two-layer GRU network. The input sequence containing noise is processed through bi-directional GRU layers (forward and backward), and the extracted features are passed to the attention mechanism module. The attention mechanism assigns greater importance to the critical information identified by BiGRU, thereby enhancing its contribution to the output and improving overall performance.

### 4.2. Sliding window smoothing of joint angle sequences

The BiGRU-Attention model can only process joint angle sequence with a fixed length. Therefore, a sliding window approach is used to scan the entire joint angle sequence. The sliding step can be set according to the video frame rate. For the *i*-th joint angle in the *j*-th frame $\theta_j^i$, the smoothed result $\hat{\theta}_j^i$ is calculated through weighted average of the corresponding joint angles extracted from $S$ sliding windows ($\hat{\theta}_1^i, \hat{\theta}_2^i, ..., \hat{\theta}_S^i$) before and after the current frame. The weights are set according to the distance $d_k$ between the adjacent *k*-th sliding window and the current window:

$$w_k = \frac{1}{d_k + \epsilon} \tag{9}$$

where $\epsilon = 0.001$ is a small constant item to avoid division by zero. The ultimate result for the *i*-th joint angle in the *j*-th frame can be expressed as:

$$\bar{\theta}_j^i = \sum_{k=1}^{S} w_k \cdot \hat{\theta}_k^i \tag{10}$$

The refined positions of keypoints and their trajectories can be obtained based on the smoothed joint angles, the stabilized base point coordinates and limb lengths.

## 5. Experimental verification

Fig. 4 shows a case of sprint. The time-course curves of joint angles at hip and knee obtained using HRNet and ViTPose++ exhibit noticeable jitters (Figs. 4(a) and 4(d)). Furthermore, significant errors occur in frames 4, 47, and 48, due to confusing the left leg with the right one (Fig. 4b). It can be observed that JAR effectively suppresses these jitters and corrects the erroneous recognition. Fig. 4(c) compares the velocity components of the right knee along *X*-axis and *Y*-axis, derived from its positions before and after the interrogated moment. The gradient calculation for velocity further aggravates the instability of raw HRNet results, leading to irregular oscillations in velocity curves. The velocities obtained by JAR show physiological consistency, conforming with the expected human motion patterns. This enhancement is critical for kinematic analysis and motion assessment. Fig. 4(d) presents a comparison of the velocity components obtained by HRNet and ViTPose++. Although unreasonable jitters are still observed in its outputs, ViTPose++ produces fewer erroneous frames. Owing to its overall superior performance, ViTPose++ is selected as the HPE model in the following performance comparisons.

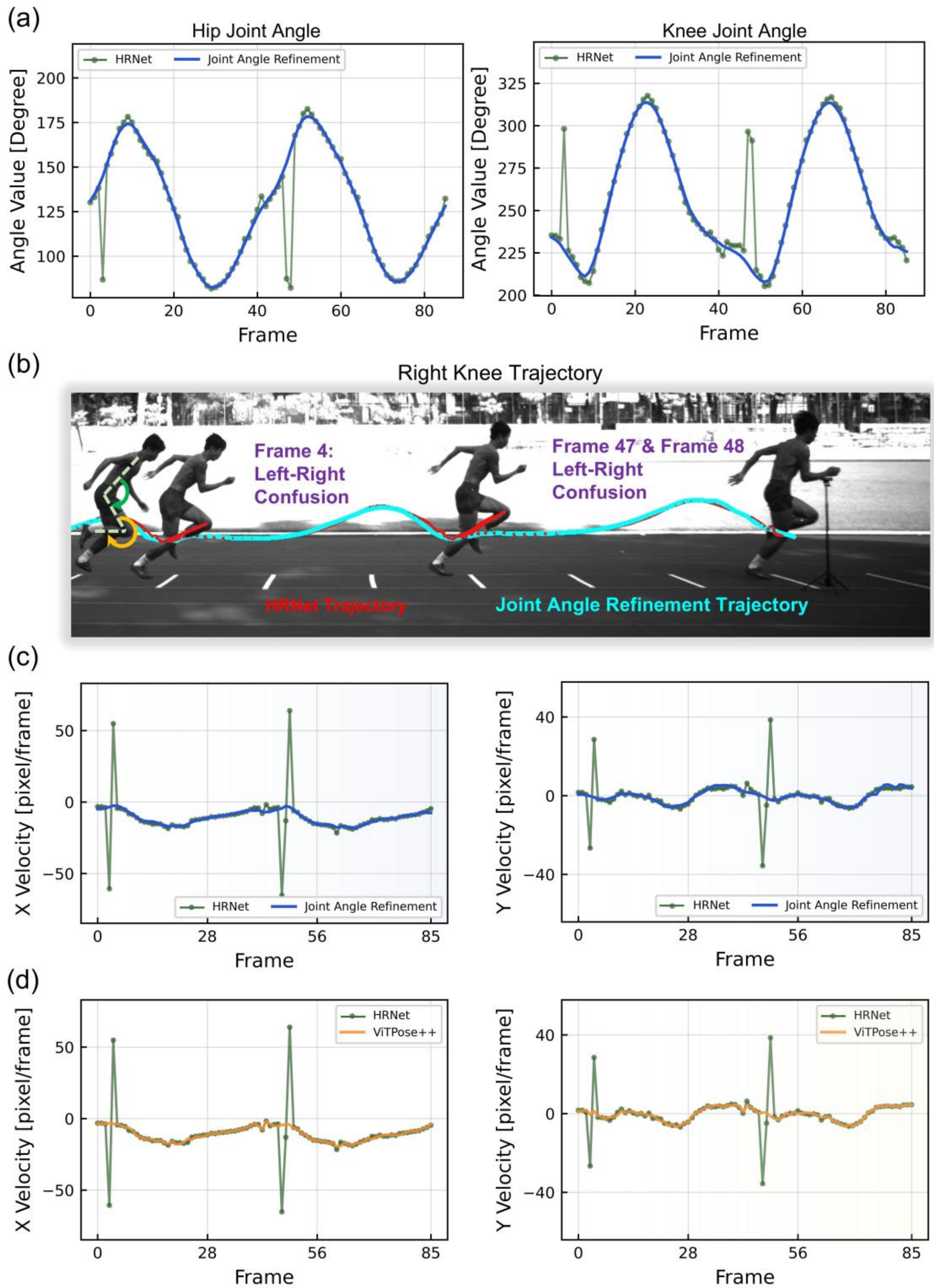

Fig. 4. Performance of JAR in a case of sprint. (a) Joint angles of hip and knee derived from raw HRNet outputs and post-processed by JAR. (b) Trajectories of right knee obtained based on the outputs of HRNet and the ones after processing of JAR. (c) Velocity components of the right knee along X-axis and Y-axis. (d) Velocity components comparison between HRNet and ViTPose++.

### 5.1. Performance comparison with SmoothNet

To demonstrate the superiority of JAR, it is compared with SmoothNet, which is regarded as the state-of-the-art HPE refinement model. The comparison is conducted in athletics and more challenging sports such as figure skating and breaking.

Fig. 5 shows the results of the two methods in a case of standing triple jump. In this activity, the postures and velocities in the preparation phase and the jumping phase are quite different. Fig. 5(a) compares directly the trajectories of right knee obtained by JAR and SmoothNet. While both methods generate physiologically plausible trajectories and correct most outliers, the curve obtained by JAR looks smoother than that by SmoothNet. In addition, there is an outlier remaining in the output of SmoothNet. Fig. 5(b) compares the evolution of $X$-component and $Y$-component separately. Besides the peak caused by the outlier, SmoothNet suffers from coordinate drift (especially along $Y$-axis in this case) when the size of smoothing window is not appropriately chosen (e.g., 32 or 64 frames). Moreover, the reliability of its results considerably declines near the outlier. In contrast, JAR maintains consistent smoothing performance across motion phases with little sensitivity to window sizes. Fig. 5(c) shows the velocities of the right knee obtained by the two methods. Compared with the smooth curves obtained by JAR, numerous sharp fluctuations are clearly visible in the results of SmoothNet. Some of them have considerable amplitudes. These spikes undoubtedly introduce disturbance to subsequent quantitative analyses.

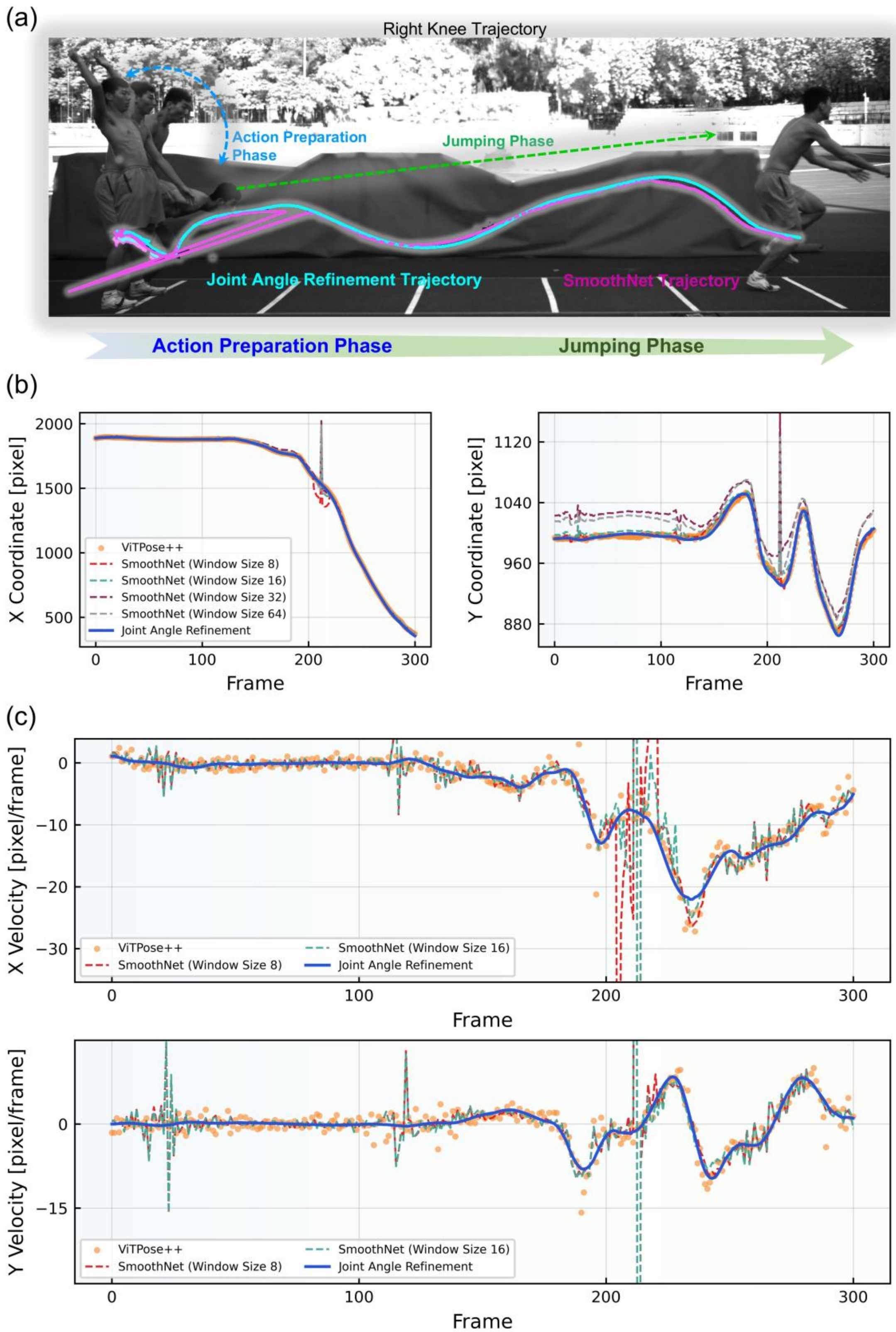

Fig. 5. Comparison of refinement performance between JAR and SmoothNet in a case of standing triple jump. (a) Trajectories of right knee obtained by JAR and SmoothNet. (b) Time-course curves of X-component and Y-component at right knee obtained by the two methods. (c) Velocity components of right knee along X-axis and Y-axis obtained by the two methods.

Table 1 summarizes the outlier correction rates reached by the two methods for all the samples in the two cases, i.e., standing triple jump and sprint. JAR achieves correction rates of 95.61% and 100%, substantially outperforming SmoothNet (53.51% and 45.45%). Specifically, SmoothNet corrected 61 out of 114 erroneous frames in standing triple jump, while JAR corrected 109 frames. In sprint, JAR corrects all 110 erroneous frames, but SmoothNet corrected only 50 of them. The overall correction rate of JAR is 97.77%, which is nearly twice that of SmoothNet.

**Table 1.** Performance of outlier correction by jar and SmoothNet in athletic activities

| Activity | Image series [group] | Image [frame] | Erroneous result [frame] | Corrected image [frame] | | Outlier correction rate |
|---|---|---|---|---|---|---|
| | | | | SmoothNet | JAR | |
| Standing triple jump | 21 | 8005 | 114 | 61 | 109 | 95.61% |
| Sprint | 20 | 8600 | 110 | 50 | 110 | 100% |
| Overall | 41 | 16605 | 224 | 111 | 219 | 97.77% |

Fig. 6 compares the refinement results of the two methods to capture the flying spin of a figure skater. This scenario presents technical challenges owing to: (i) Rapid alternating positions of the left and the right lower-limb joints; (ii) Visual indistinctness of the left and the right low-limb joints due to dark-colored pants. Figs. 6(a) and 6(b) show the trajectories obtained by the two methods along with raw ViTPose++ outputs. ViTPose++ outputs are severely disturbed by left-right confusion and inter-frame inconsistency. In the trajectory refined by SmoothNet, most of outliers are eliminated but the turns and shifts look somewhat unnatural. In contrast, JAR gives more smooth and stable motion trajectories that precisely capture the biomechanical coherency of this maneuver. Figs 6(c) and 6(d) show the calculated velocity components along *X*-axis and *Y*-axis, respectively. The results of SmoothNet exhibit high-frequency fluctuations and contain several anomalous spikes, whereas the results of JAR are smoother and more continuous, indicating its superiority in suppressing inter-frame inconsistency of keypoint positions.

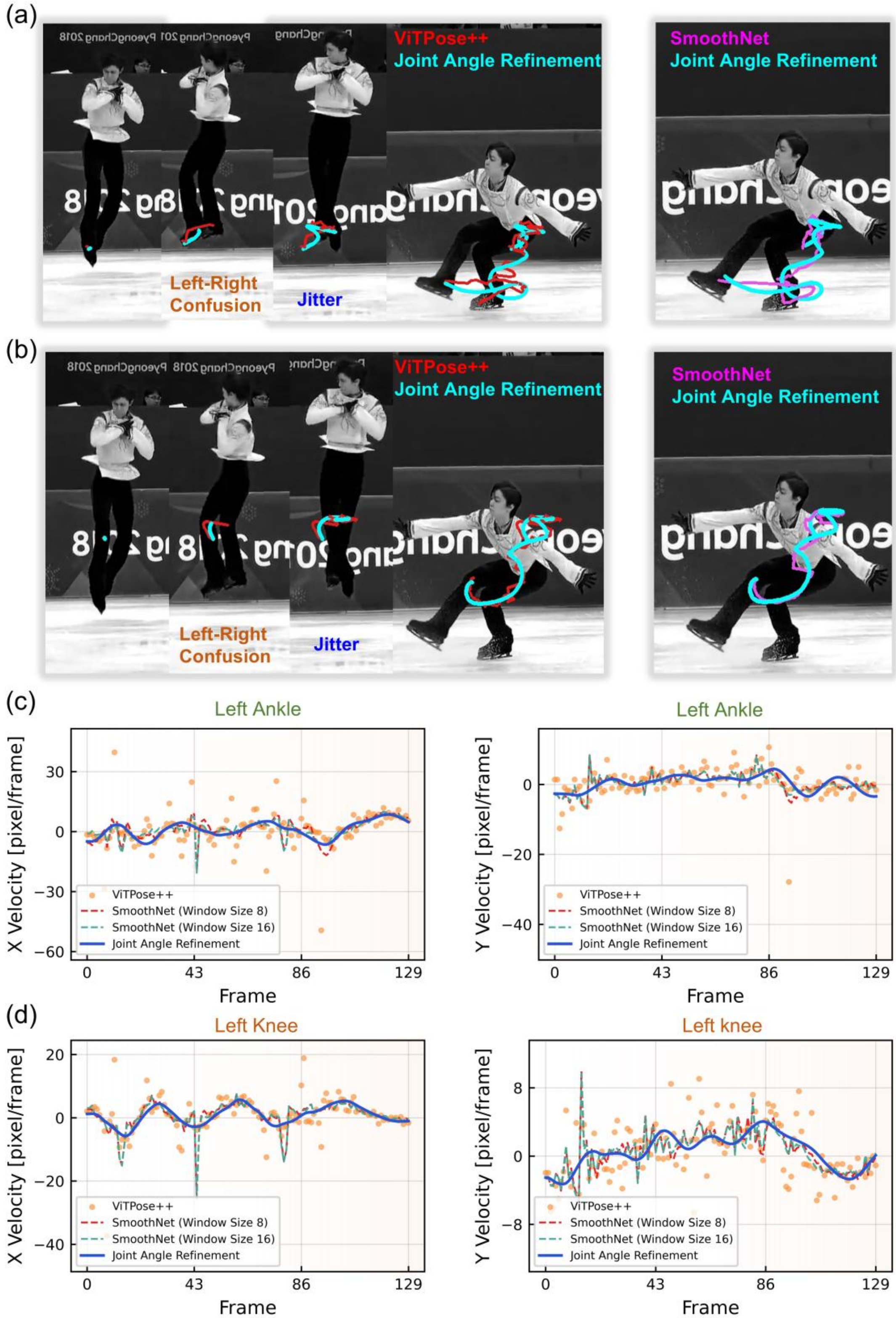

Fig. 6. Comparative analysis of refinement performance between JAR and SmoothNet in a case of figure skating. Trajectories of (a) the left ankle and (b) the left knee obtained by the two methods along with raw ViTPose++ outputs are compared. The difference becomes more distinguishable when comparing velocity components of the left knee along (c) X-axis and (d) Y-axis. To demonstrate the process from left side to right, the images are mirrored.

Fig. 7 shows a case of handstand in breaking. The challenges come from considerable limb-folding and intermittent motions. Fig. 7(a) compares the trajectories of the left ankle obtained by the two methods. JAR gives the smoothest curve among the three, which well conforms with the biomechanical features of the maneuver. The gap seems clearer in Figs. 7(b) and 7(c), in which the position variations of the left ankle and the velocity variations of the left shoulder are compared. The refinement of SmoothNet is considerably interfered by the unusual mode of this activity. Many abrupt fluctuations are found in the time-course curves of *X*-coordinates using smoothing windows of various sizes (the left subfigure in Fig. 7(b)). The left subfigure in Fig. 7(c) also shows numerous spikes in the curves of velocity. Moreover, substantial drift from the correct values is clearly visible in the time-course curves of *Y*-coordinates (the right subfigure in Fig. 7(b)). The relatively poor performance could be ascribed to the scarcity of samples in current datasets, as Breaking is a new competitive event added to the Olympic Games in 2024. However, the results of JAR accurately catch the motions and reflect the characteristics of this maneuver. The *X*-component of velocity varies in a manner of high frequency and small amplitude, because during the handstand, player tries to keep the axis of body vertical but there is inevitable slight swing. The *Y*-component of velocity also stays at a steady base line, with occasional rapid and continuous changes due to folding of leg.

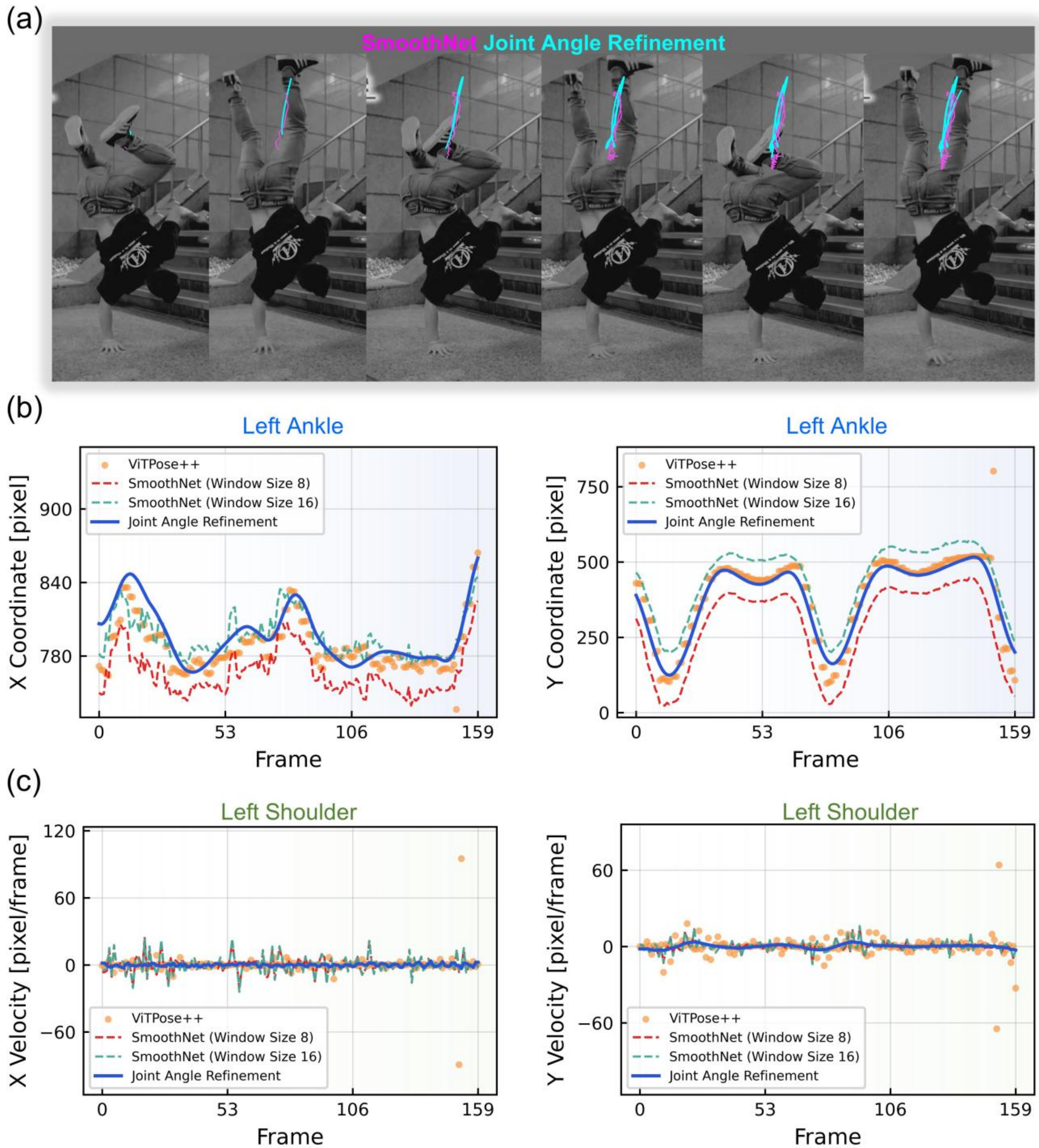

Fig. 7. Comparison of refinement performance between JAR and SmoothNet in a case of breaking. (a) Trajectory of the left ankle obtained by JAR and SmoothNet. (b) Coordinates of the left ankle along X-axis and Y-axis. (c) Velocity components of the left shoulder along X-axis and Y-axis.

### 5.2. Selection of sequence-to-sequence models

Deep learning based sequence-to-sequence models have been widely used in denoising of time series data [54–58], due to their superior adaptability over traditional filters to signal variations. To reconstruct a smooth sequence of pose conforming with biomechanical motion patterns from a noisy input sequence, the deep learning model need not only capture the correlation between the local features in a short time interval but also extract global features in whole process. This study systematically evaluates the performance of ten mainstream sequence-to-sequence architectures in HPE refinement, including temporal convolutional network (TCN) [59], LSTM [60], bi-directional LSTM (BiLSTM) [61] [62], BiGRU [63], and their attention-augmented variants (TCN-Attention, BiLSTM-Attention and BiGRU-Attention), as well as two recently-established state space models, i.e., Mamba1 (proposed in 2023) [64] and its subsequent advancement Mamba2 (proposed in 2024) [65].

Fig. 8 compares the performance of the ten models for refining joint angle variations of knee and hip over a complete motion cycle in testing set. The "ground truth" sequences are constructed using an 8th-order Fourier series with randomly superimposed jitters and outliers. It can be observed that all the ten models, trained with our dataset, are able to smooth the jerky curves and correct most of the outliers. However, BiGRU and BiGRU-attention give the outputs closer to the "ground truth", especially near the turning points of joint angle (i.e., the peaks and troughs), indicating their stronger ability to catch motion variations (see insets in Figs. 8(a) and 8(b)). To quantitatively evaluate the refinement performance, the mean squared error (*MSE*) is adopted as the primary metric:

$$MSE = \frac{1}{N}\sum_{i=1}^{N}\left(\hat{\theta}_i - \theta_i\right)^2 \tag{11}$$

where $N$ denotes the total number of frames, $\theta_i$ represents the "ground truth" of joint angles, $\hat{\theta}_i$ indicates the refined values. Fig. 8(c) compares the *MSE* achieved by the ten models. Among the models without attention mechanisms, the models based on recurrent neural network (RNN), i.e., LSTM, BiLSTM, BiGRU, significantly outperform the models based on CNN. Additionally, bi-directional mechanism is found helpful to enhancing the adaptability of models to temporal variation. The *MSE* achieved by BiLSTM is approximately 44% lower than that by LSTM. BiGRU demonstrates better accuracy than BiLSTM, with *MSE* 46% lower. Attention mechanism also contributes considerably to improving the accuracy of BiGRU and BiLSTM. The *MSE* achieved by BiLSTM-Attention and BiGRU-Attention are reduced by about 66% and 44%, respectively. Mamba models, though theoretically strengthen the modeling ability of signal correlation over long time ranges, their performance in this case is relatively weak.

In this study, BiGRU-Attention is selected as the optimum option due to its balanced performance, robustness, and computational efficiency. It is noteworthy that other RNN-based models, benefiting from the inherent robustness of joint angle-based description of human pose and the high-quality training dataset constructed using Fourier series approximation, can achieve acceptable refinement performance even without attention mechanisms.

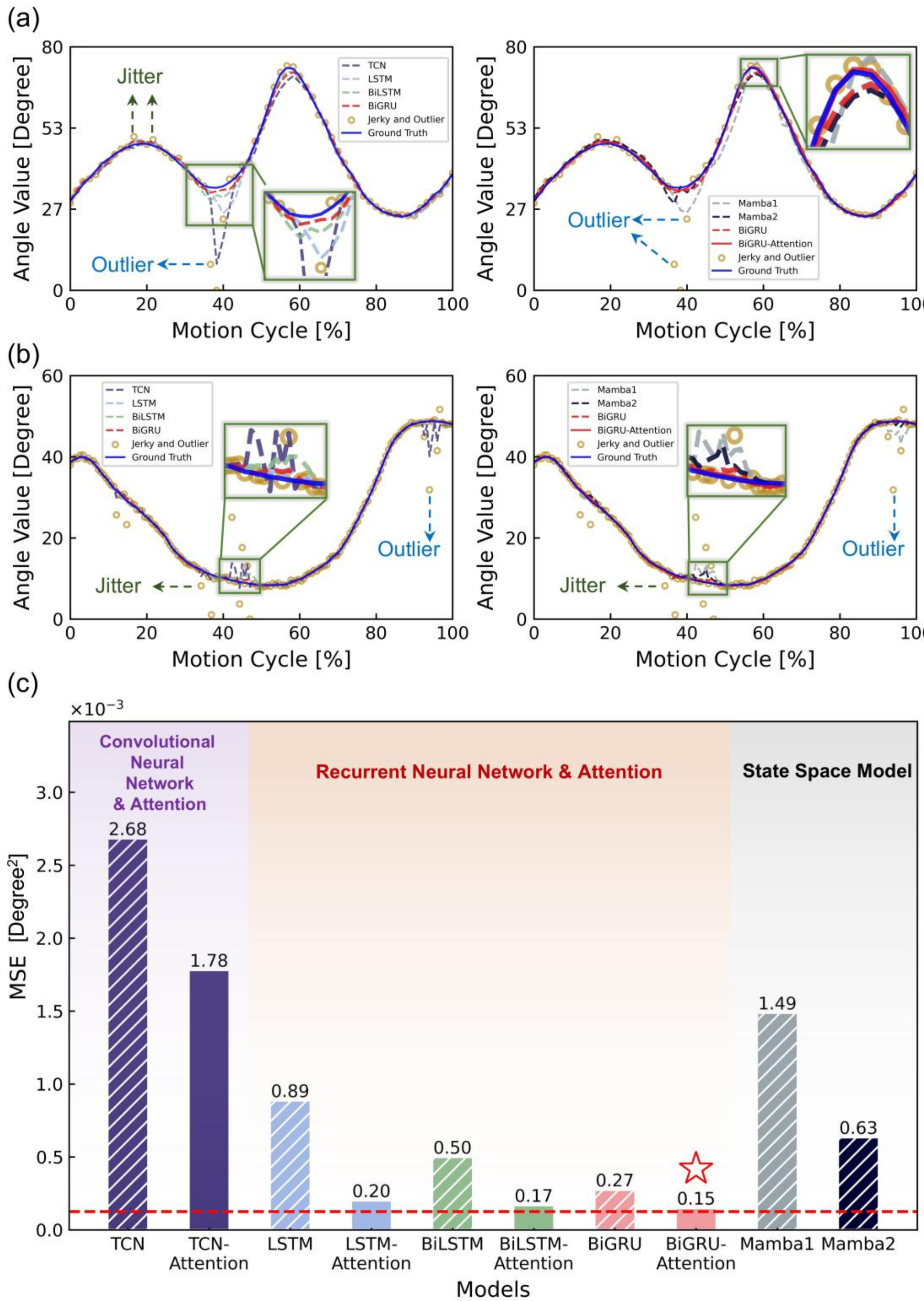

Fig. 8. Performance of ten mainstream deep learning-based sequence-to-sequence models for refining noisy temporal variation of joint angles at (a) knee and (b) hip. (c) Comparison of mean squared error achieved by the models.

### 5.3. Correction of video datasets

The proposed method can also be used to rectify the annotations in existing HPE video datasets,

eliminating the inter-frame inconsistencies caused by manual annotating. Fig. 9 shows an typical example, in which JAR is employed to correct the annotations in the PoseTrack dataset [13]. In Fig. 9(a), the trajectories of shoulder, hip and ankle of the female dancer constructed according to the original annotations contain numerous jitters. After the processing of JAR, the trajectories become significantly smoother. Figs. 9(b) and 9(c) compare the curves of velocity components at the left shoulder and ankle calculated before and after annotation correction. There are a lot of abrupt fluctuations in the velocity curves obtained using the original annotations, clearly showing the adverse influence of human-induced bias in manual annotating. The biomechanical implausibility is substantially mitigated by JAR. Furthermore, JAR contributes not only to the smoothness of keypoint trajectories but also to the localization accuracy of keypoints. The refined results, conforming better with human motion, enhances the reliability of training datasets and reduces the potential misinformation to models.

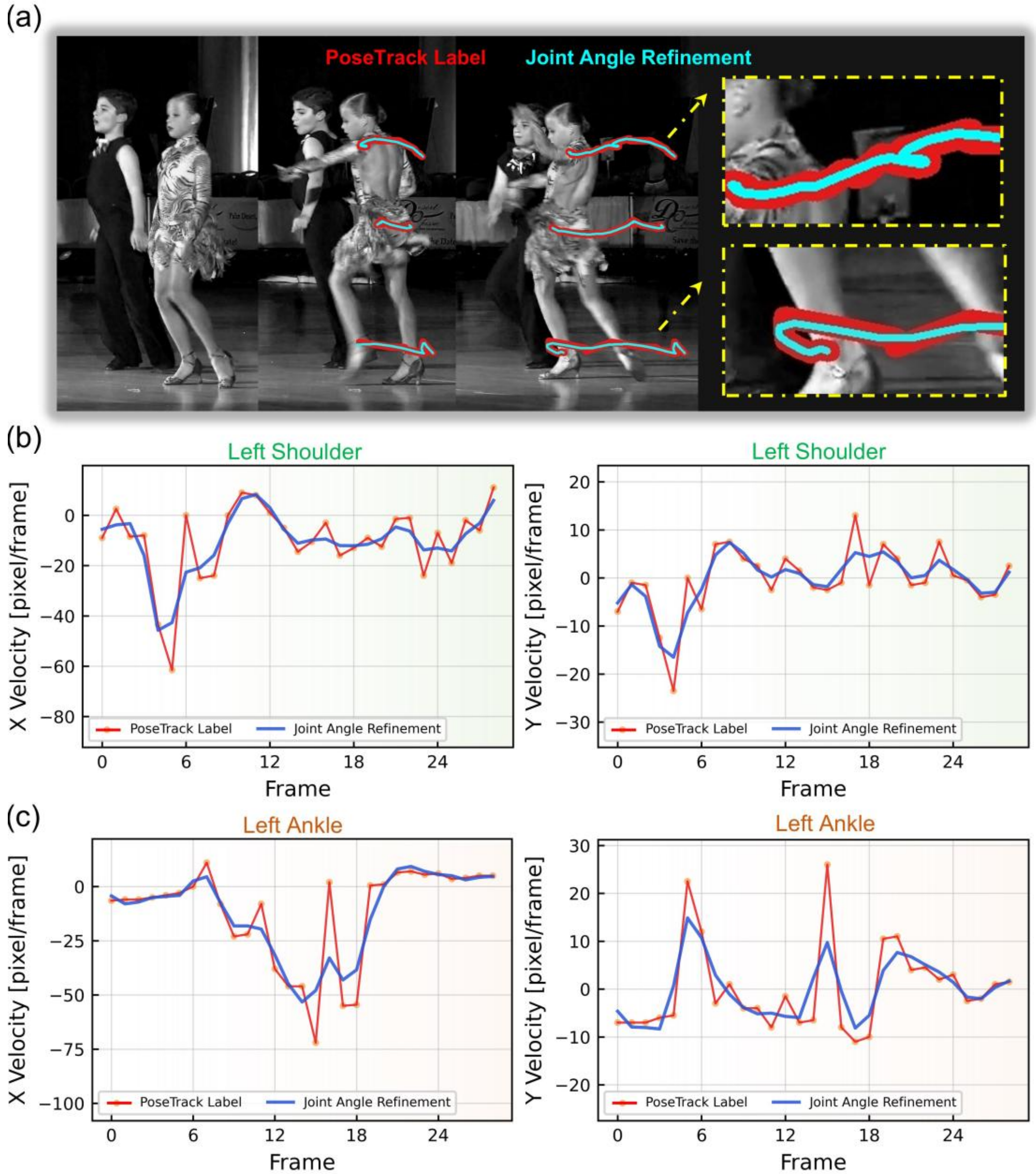

Fig. 9. Correction of annotations for a sample in the PoseTrack dataset using JAR. (a) Trajectories of shoulder, hip and ankle obtained by original annotations and JAR results. (b) Velocity components of left shoulder and (c) left ankle along X-axis and Y-axis calculated according to the original annotations and JAR results.

## 6. Conclusion

The study indicates that the capability of current deep learning-based models for HPE refinement of kinematic poses is limited largely by the datasets with insufficient samples and inaccurate annotations. Our method demonstrates how to establish robust description of human poses during activities based on joint angles and to generate highly reliable training dataset through Fourier series-based approximation of spatiotemporal variation of joint angles. The high-quality dataset significantly enhances the performance of RNN-based sequence-to-sequence models to eliminate jitters and outliers in reconstruction of keypoint trajectories. It also endows the refinement model with outstanding generalization ability to tackle the challenging cases in sports. Furthermore, the method shows great potential to rectify the annotations in existing video datasets, which could prompt substantial performance improvement of HPE technology for analysis of human motion.

## Acknowledgments

The study was financially supported by the National Natural Science Foundation of China (Grant Nos. 12232017, 12472179, and 12572432) and Natural Science Foundation of Guangdong Province (Grant Nos. 2024A1515011076, 2025A1515011999, and 2026A1515011088).